\documentclass[letterpaper, 10 pt, conference]{ieeeconf}
\usepackage{amsmath}
\usepackage{mathtools}
\usepackage{algorithm}
\usepackage[noend]{algpseudocode}
\usepackage{hyperref}
\usepackage{multirow}

\usepackage[dvipsnames]{xcolor}

\usepackage[most]{tcolorbox}
\newtcolorbox{highlighted}{colback=yellow,coltext=red,breakable}

\usepackage{subcaption}
\usepackage{kantlipsum} 
\usepackage{mwe} 
\usepackage[squaren, Gray, cdot]{SIunits}
\usepackage{hhline}

\DeclareCaptionLabelSeparator{periodspace}{.\quad}
\IEEEoverridecommandlockouts                              

\title{\LARGE \bf Decentralized Multi-Agent Planning for Multirotors:  \\ a Fully Online and Communication Latency Robust Approach}

\author{Charbel Toumieh 
\thanks{The author is an independent researcher
(e-mail: \url{charbel.toumieh@gmail.com})}%
}
\begin{document}

\setlength{\topmargin}{-17pt}

\maketitle
\thispagestyle{empty}
\pagestyle{empty}

\begin{abstract}
There are many industrial, commercial and social applications for multi-agent planning for multirotors such as autonomous agriculture, infrastructure inspection and search and rescue. Thus, improving on the state-of-the-art of multi-agent planning to make it a viable real-world solution is of great benefit. In this work, we propose a new method for multi-agent planning in a static environment that improves our previous work by making it fully online as well as robust to communication latency. 
The proposed framework generates  a global path and a Safe Corridor to avoid static obstacles in an online fashion (generated offline in our previous work). It then generates a time-aware Safe Corridor which takes into account the future positions of other agents to avoid intra-agent collisions. The time-aware Safe Corridor is given  with a local reference trajectory to an MIQP (Mixed-Integer Quadratic Problem)/MPC (Model Predictive Control) solver that outputs a safe and optimal trajectory. The planning frequency is adapted to account for communication delays. The proposed method is fully online, real-time, decentralized, and synchronous. It is compared to 3 recent state-of-the-art methods in simulations. It outperforms all methods in robustness and safety as well as flight time. It also outperforms the only other state-of-the-art latency robust method in computation time.
\end{abstract}
\textbf{ }

 \textbf{video}: \url{https://youtu.be/eKwYNU1Q0wY}

\section{INTRODUCTION}
\subsection{Problem statement}
Multi-agent planning has been gaining in popularity in the research community due to recent advances. These advances are making it a viable solution to many commercial, industrial, and military applications. There are multiple challenges that face a multi-agent planning framework such as the problem of synchronizing agents for synchronous planning methods and dealing with communication latency. It is the purpose of this paper to extend upon our previous state-of-the-art work \cite{toumieh2022multi} that outperformed other state-of-the-art methods in computation efficiency, trajectory speed, and smoothness in a cluttered environment. We provide a new approach derived from \cite{toumieh2022multi} that is fully online and robust to arbitrary communication latency. We also study the effect of communication latency on the overall performance of our planner and compare it with other state-of-the-art methods. 

\subsection{Related work} \label{sect:related_works}
\subsubsection{Multi-agent planning for multirotors}
In {\cite{Hoenig2018}}, the authors present a centralized multi-agent planning framework that uses time-aware Safe Corridors. The method has 3 sequential steps: roadmap generation, then discrete planning, and finally continuous refinement. The approach presented by the authors is centralized although some steps can be decentralized. While the computation time is not suitable for online high-speed planning and replanning, the method used served as an inspiration for many subsequent methods in the state-of-the-art. Such methods include \cite{Park2022} and \cite{toumieh2022multi} which in turn served as an inspiration for the work presented in this paper.

Buffered Voronoi Cells have been used by multiple works \cite{Zhu2019bvc}, \cite{Zhou2017} for multi-agent collision avoidance but do not account for static obstacles. Other approaches \cite{Luis2019dmpc} use separating hyperplanes to avoid collisions between agents and model static obstacles in the form of ellipsoid constraints in a decentralized MPC formulation. The generation of ellipsoid representation of the environment is not trivial and is not addressed by the authors of \cite{Luis2019dmpc}.

MADER, an asynchronous multi-agent planning framework has been proposed in \cite{tordesillas2020mader}. The method allows for avoiding static, and dynamic obstacles, as well as other planning agents. The authors combine a search-based approach with an optimization approach, where the output of the search-based approach is taken as initialization for the optimization problem. This choice was made since the optimization problem defined by the authors is non-convex and requires a good initial guess.

EGO-Swarm was proposed in \cite{Zhou2021EGOSwarmAF} as an asynchronous and decentralized trajectory planner. It requires each planning agent to broadcast its generated trajectory at a fixed frequency. When each agent receives the trajectories of other agents, it proceeds immediately to do a collision check. While the approach has been demonstrated in real-world experiments, it still suffers from collisions due to communication delays between agents.

In a similar fashion to \cite{Hoenig2018}, the authors of \cite{Park2022} present a distributed and online trajectory generation framework for multi-agent quadrotor systems using time-aware Safe Corridors (or Linear Safe Corridors). The environment representation used by the authors is an octomap {\cite{hornung2013octomap}}. The Safe Corridor used to generate the time-aware Safe Corridor contains only one polyhedron which leads to slow and conservative trajectories.

In \cite{soria2021distributed}, a decentralized model predictive control approach is used for collision avoidance and cohesive flight. The obstacles are described as mathematical functions (cylinders, paraboloids ...) in order to include them in the decentralized MPC formulation as constraints. It is however not trivial to describe an arbitrary cluttered environment through continuous mathematical functions that are easy to add as constraints to an MPC formulation.

Finally, in our previous work \cite{toumieh2022multi}, we proposed a decentralized and synchronous planning framework that is inspired by \cite{Hoenig2018}. The approach takes into account static obstacles using Safe Corridors (generated from a voxel grid representation \cite{toumieh2020mapping}). Safe Corridors are then augmented to time-aware Safe Corridors to avoid intra-agent collisions. The proposed approach outperforms state-of-the-art methods in all performance metrics, including robustness, computation time, and trajectory speed.

\subsubsection{Latency robust multi-agent planning}
The previously cited works do not account for communication delay, or can passively handle latency up to a fixed limit \cite{toumieh2022multi}. Some multi-agent planning frameworks take into account communication delay and will be presented in this section.

In \cite{senbaslar2022asynchronous}, an asynchronous and decentralized trajectory planner is presented. The planner guarantees safety using separating hyperplanes from previous planning iterations. While the presented approach can handle communication delays, it does not account for any type of obstacles (static or dynamic), which limits its applicability to the real world.

Finally, RMADER (Robust MADER) is proposed in \cite{kondo2022robust}, which is an extension of MADER \cite{tordesillas2020mader}. They convexify the optimization problem in order to improve the computation time. However, they inherit from MADER the polyhedral representation of the obstacles in the environment. This representation is not trivial to generate and can add significant overhead to the planning framework.

\subsection{Contribution}
The main contribution of our paper is an improved decentralized and synchronous planning framework that is robust to communication latency. The proposed framework is built on our previous work \cite{toumieh2022multi} and conserves its advantages. Thus, the proposed method has low computation time and takes into account static obstacles and other planning agents.
The improvements are:
\begin{enumerate}
\item The addition of a mechanism to deal with arbitrary communication latency by dynamically adapting the planning frequency to avoid collisions and guarantee safety.
\item The integration of 2 previously offline steps in \cite{toumieh2022multi} (global path generation step and Safe Corridor generation step) to make the framework fully online and suitable for real-world applications.
\item The modification of the stalemate/deadlock resolution mechanism to guarantee safety.
\end{enumerate}
The method is tested in simulations to show the effect of communication latency on the performance of the planner. It is also compared to 3 recent works: EGO-Swarm \cite{Zhou2021EGOSwarmAF}, MADER \cite{tordesillas2020mader} and RMADER \cite{kondo2022robust} in terms of trajectory safety/performance as well as computation time.

\section{Assumptions} \label{sect:assumptions}

\begin{figure}
\centering
\includegraphics[trim={0cm 0 0 -1cm},clip,width=1\linewidth]{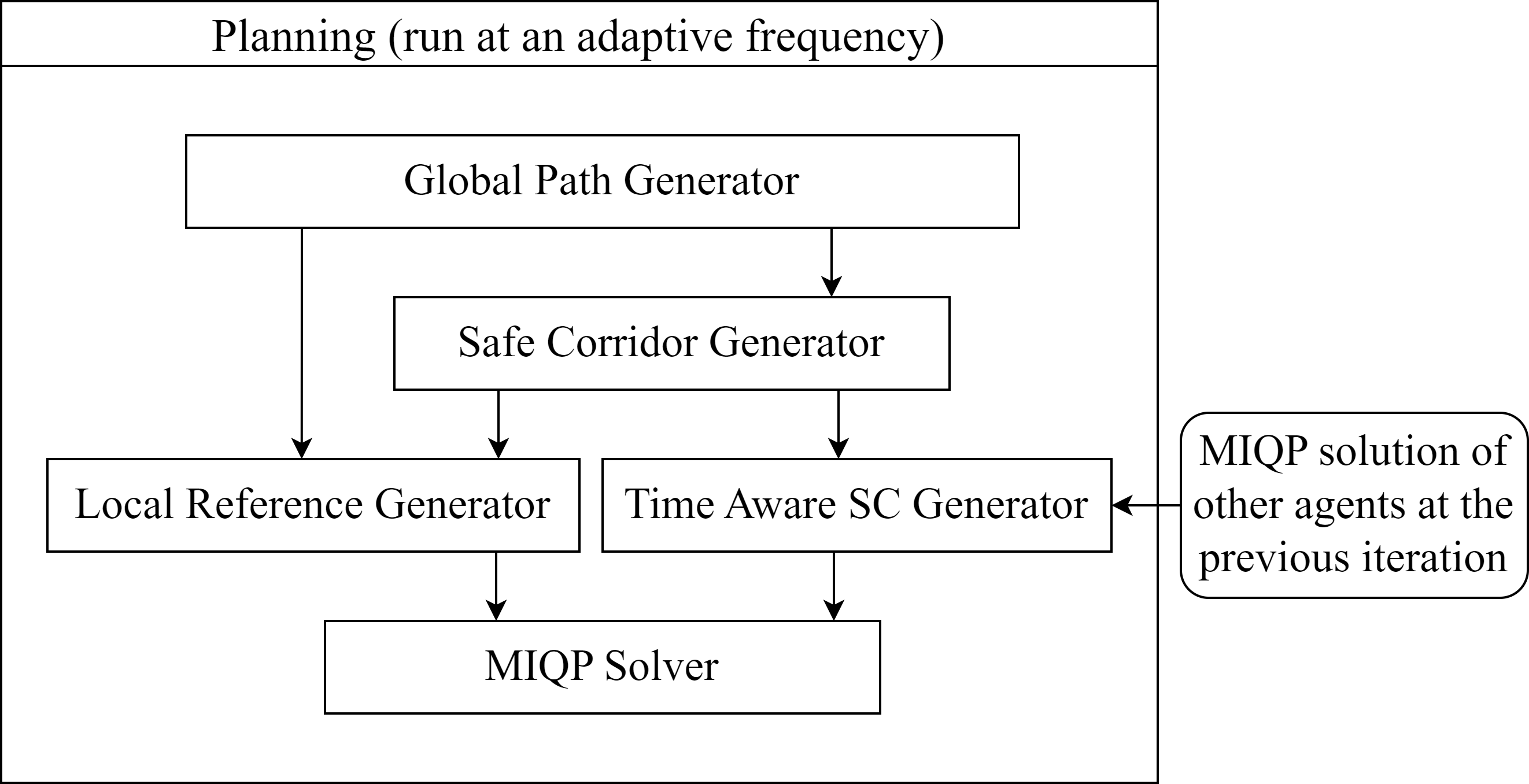}
\caption{We show the global pipeline of the planning framework of a single planning agent. It is run in a loop at a varying/adaptive frequency.}
\label{fig:diagram}
\end{figure}

We assume perfect control (the controller executes the generated trajectory perfectly) and perfect localization (each agent can localize itself and other agents at any moment to an arbitrary accuracy). These assumptions are made by all of the previously cited state-of-the-art methods. In addition to these assumptions, we assume that the clocks of the agents are synchronized. We assume 2 cases: 
\begin{enumerate}
\item We can synchronize all agents at the beginning of a given mission.
\item If an agent (not synchronized) is getting close to a cluster of other synchronized agents, we assume the range of communication is big enough so that the agent can synchronize its clock with the cluster before getting close enough for collision avoidance.
\end{enumerate}
Furthermore, we assume symmetric behavior of the communication: if there is a latency in the delivery of a message from agent $i$ to agent $j$ in a given planning iteration/period, the same latency happens when agent $j$ is trying to deliver a message to agent $i$.

\begin{figure*}
\begin{subfigure}{0.5\textwidth}
\centering
\includegraphics[trim={0cm 0cm 0cm 0cm},clip,width=0.8\linewidth]{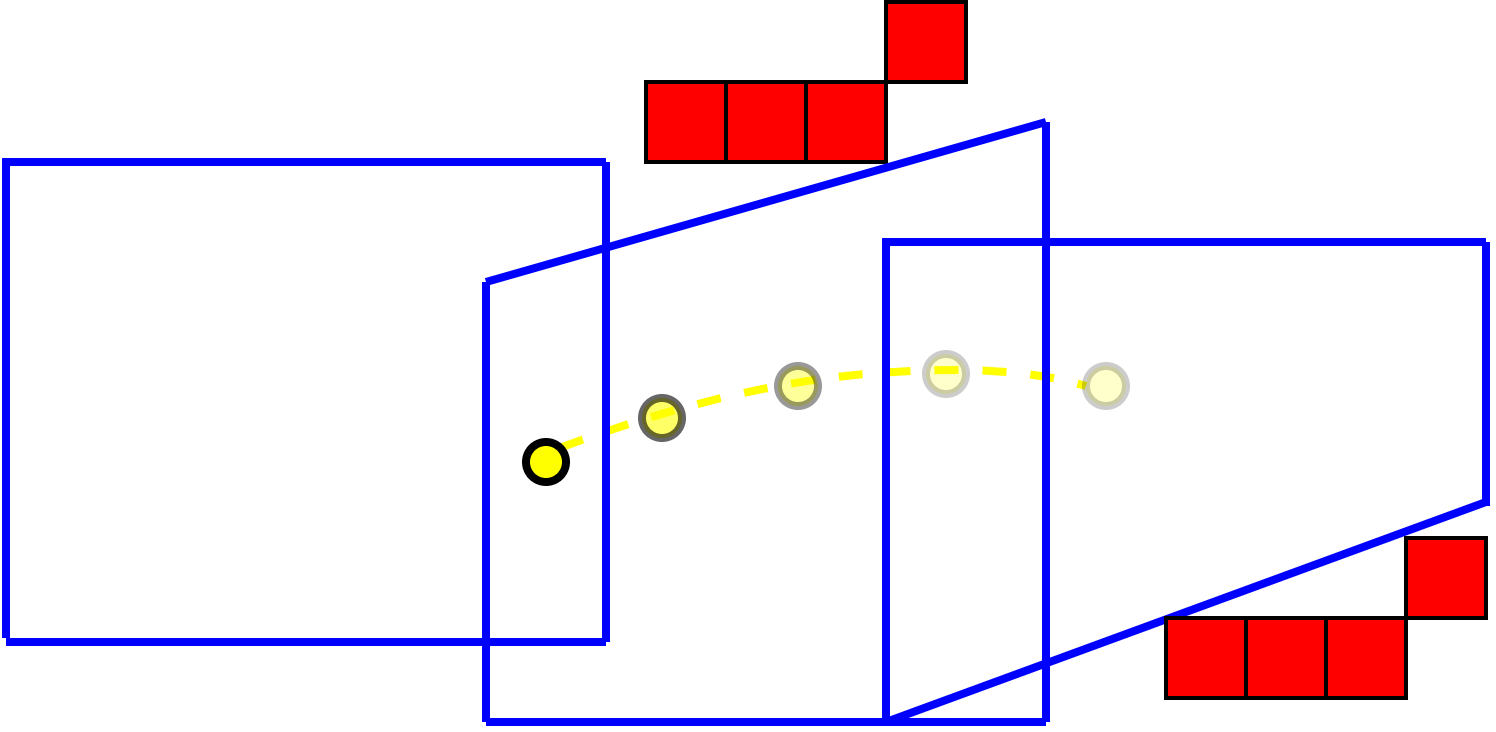}
\caption{Safe Corridor at iteration $k$.}
\label{fig:sc_evo_1}
\end{subfigure}
\begin{subfigure}{0.5\textwidth}
\centering
\includegraphics[trim={0cm 0cm 0cm 0cm},clip,width=0.9\linewidth]{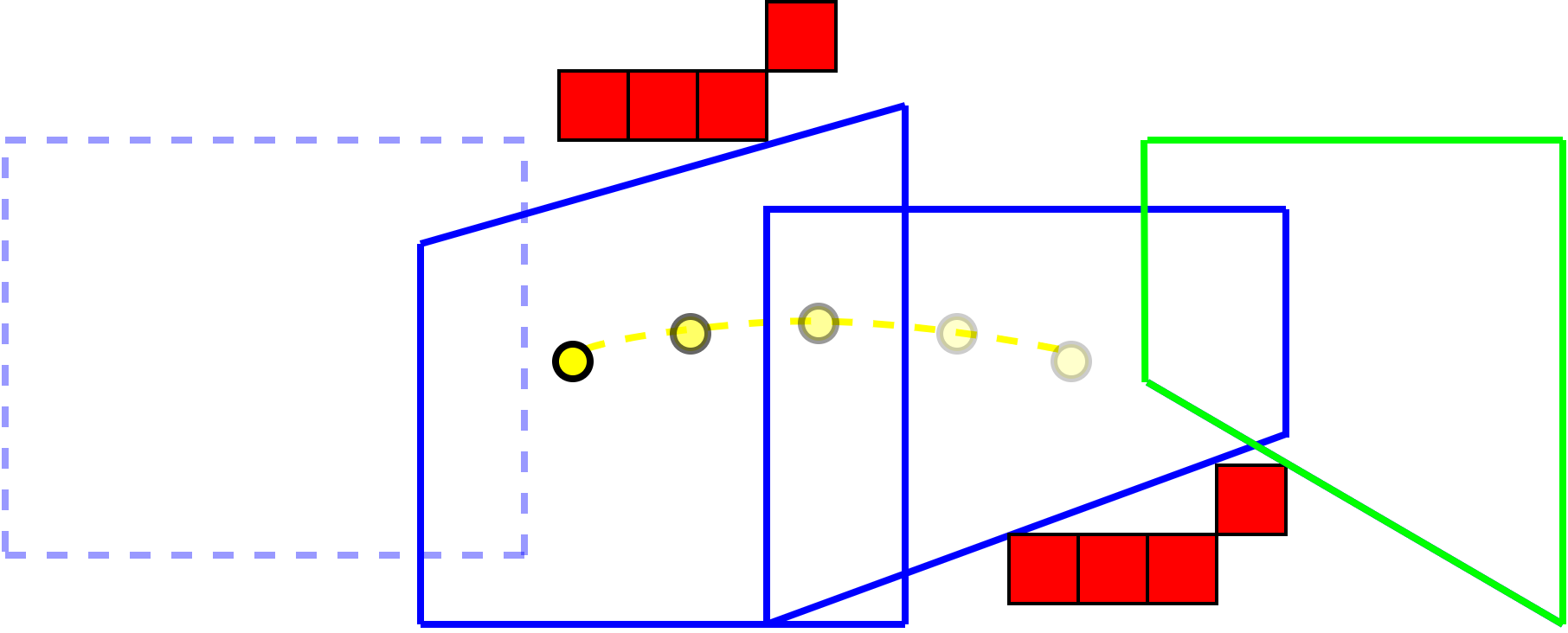}
\caption{Safe Corridor at iteration $k+1$.}
\label{fig:sc_evo_2}
\end{subfigure}

\caption{The obstacles are shown in \textbf{red}. The predicted positions of the agent are shown as \textbf{yellow} circles (MPC trajectory). They get increasingly transparent as we move forward in time. At iteration $k$ (Fig. \ref{fig:sc_evo_1}), all polyhedra (in \textbf{blue}) contain at least one point of the MPC trajectory. At the next iteration $k+1$ (Fig. \ref{fig:sc_evo_2}), the first position of the MPC trajectory moves out of the first polyhedron (in \textbf{dashed blue} lines). Thus, we remove it from the Safe Corridor and generate another polyhedron (in \textbf{green}) using the global path. The new polyhedron is added to the Safe Corridor.}
\label{fig:update_sc}
\end{figure*}

\section{The planner} \label{sect:the_method}
Our planner is run concurrently on each agent in a swarm. The dynamical model of each agent is the same as presented in \cite{toumieh2022multi}. We use a voxel grid representation of the environment, which can be trivially and efficiently generated \cite{toumieh2020mapping}. Each agent has a voxel grid that is of fixed size and that moves with the agent such that the agent is always at its center. This voxel grid is used for global path finding and Safe Corridor generation. The clocks of the agents are synchronized. 

In \cite{toumieh2022multi}, the planning is divided into 2 stages: an offline stage for global path finding and Safe Corridor generation; then an online stage where the time-aware Safe Corridors and the dynamically feasible trajectory are generated. In the planner proposed in this paper, the offline stage is now integrated into the online planning stage so the whole planning/replanning framework is run online. This makes it suitable for real-world deployment and missions such as exploration. The steps of the proposed planner are (Fig. \ref{fig:diagram}):
\begin{enumerate}
    \item Generate a global path (Sect. \ref{sect:global_path}).
    \item Generate a Safe Corridor (Sect. \ref{sect:sc})
    \item Generate a time-aware Safe Corridor (Sect. \ref{sect:ta_sc}).
    \item Generate a local reference trajectory (Sect. \ref{sect:loc_ref}).
   \item Solve the Mixed-Integer Quadratic Program (MIQP)/Model Predictive Control (MPC) problem to generate a locally optimal trajectory (Sect. \ref{sect:form}).
\end{enumerate}

In the first step, we generate a global path from the position of the agent to the goal position. This path avoids all static obstacles and is used to generate the Safe Corridor and to generate the local reference trajectory.
In the second step, we generate a Safe Corridor (a series of overlapping convex polyhedra) that covers only the free space in the environment. These convex polyhedra are used as linear constraints in an optimization formulation to constrain the trajectory to the free space and avoid collisions with static obstacles.
In the third step, we use the recently generated trajectories of the agents and the Safe Corridor to generate time-aware Safe Corridors. This allows the agents to avoid intra-agent collisions.
In the fourth step, we sample the global path at a given velocity to generate a local reference trajectory that the dynamically feasible trajectory tries to follow as closely as possible.
In the fifth and final step, we generate the dynamically feasible trajectory to be executed by the agent. It is generated by solving an optimization problem that takes time-aware Safe Corridors and a local reference trajectory and guarantees that there are no collisions of any nature (intra-agent or static obstacles) while the agent moves closer to its goal.

These steps were run sequentially and periodically at a fixed frequency in our previous work \cite{toumieh2022multi}. However, in this work, we vary the planning frequency to account for communication latency. As in \cite{toumieh2022multi}, each agent broadcasts its planned trajectory at the end of the planning iteration so that other agents can know it. In addition to the planned trajectory, we also broadcast the times we started and finished generating the trajectory so that other agents can estimate the communication latency (not done in \cite{toumieh2022multi} - more details in Sect. \ref{sect:com_latency}). We briefly explain each step in this section while focusing more on the steps where changes were made with respect to \cite{toumieh2022multi}.

\subsection{Generate a global path} \label{sect:global_path}
In this step, a global path is generated connecting the current position of the agent to the desired final position using the local voxel grid. The occupied voxels in the voxel grid are inflated by each agent's size before feeding the grid to the path planning algorithm. In case the goal position is outside the local voxel grid of the agent, we choose an intermediate goal in the grid as presented in \cite{toumieh2020planning}. The main idea is to draw a line connecting the position of the agent to the goal and get the intersection with the borders of the voxel grid. This intersection is a voxel and is set as an intermediate goal. We also clear/set to \textbf{free} all the border voxels of the voxel grid to help the agent find a path to the intermediate goal in extremely cluttered environments.

At each iteration, the starting point for the global path search is the last point in the local reference trajectory generated in the previous planning iteration (Sect. \ref{sect:loc_ref}). The local reference trajectory is then connected to the path found through the global search to generate the final global path used in the subsequent sections (for generating the local reference trajectory of the current iteration).

We use JPS (Jump Point Search) \cite{harabor2011online} and DMP (Distance Map Planner) for path planning. JPS employs pruning techniques on the A* algorithm to potentially speed up the generation time by an order of magnitude. DMP uses artificial potential fields to push the path generated by JPS away from obstacles. This adds an additional margin of safety and improves the trajectory generated in the last step (MIQP optimization output) in terms of speed and smoothness (see \cite{toumieh2020planning} for more details).
\begin{figure*}
\begin{subfigure}{0.33\textwidth}
\centering
\includegraphics[trim={0cm 0cm 0cm 0cm},clip,width=0.9\linewidth]{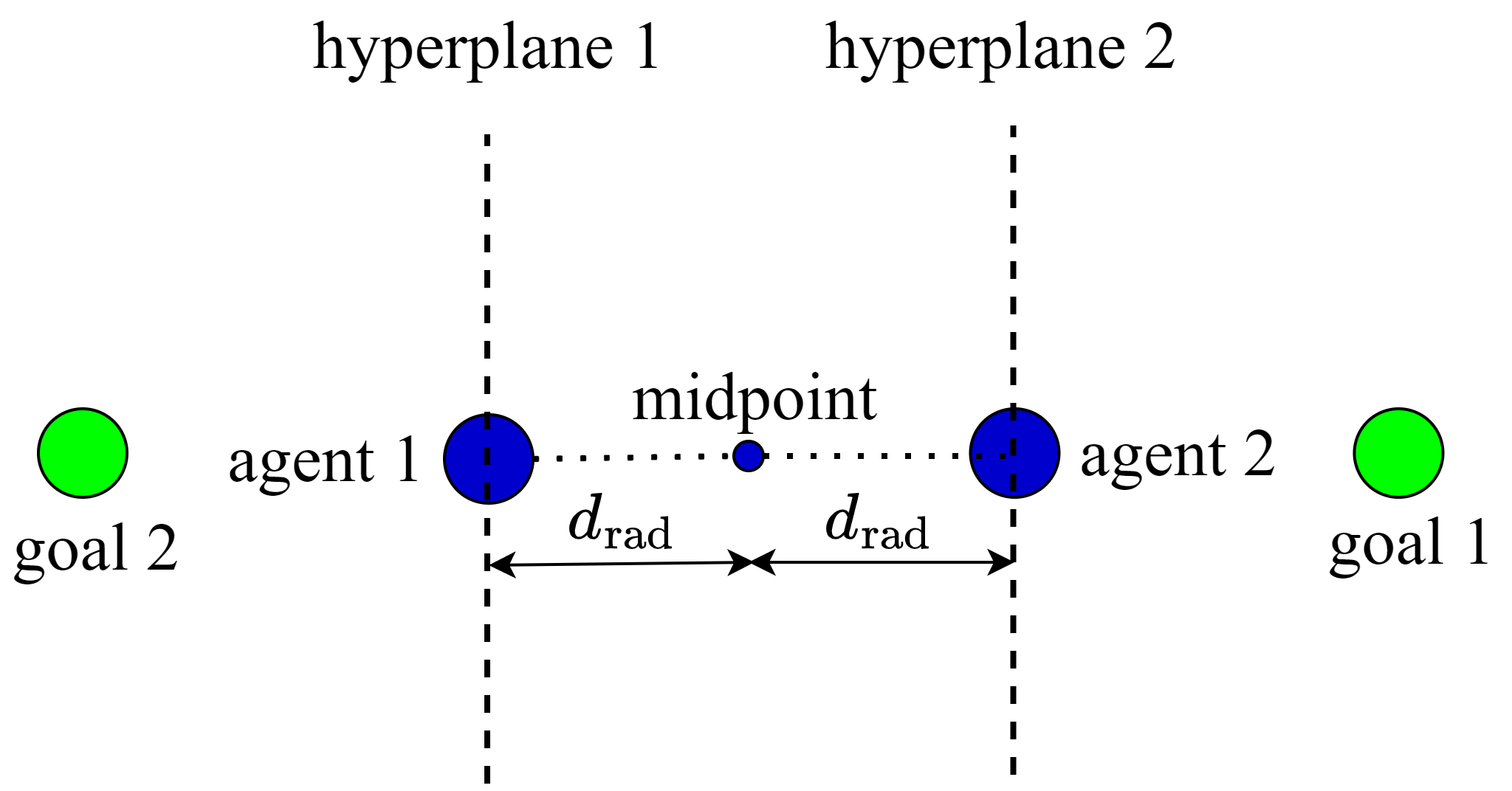}
\caption{Stalemate caused by a symmetrical position.}
\label{fig:stalemate_1}
\end{subfigure}
\begin{subfigure}{0.33\textwidth}
\centering
\includegraphics[trim={0cm 0cm 0cm 0cm},clip,width=0.9\linewidth]{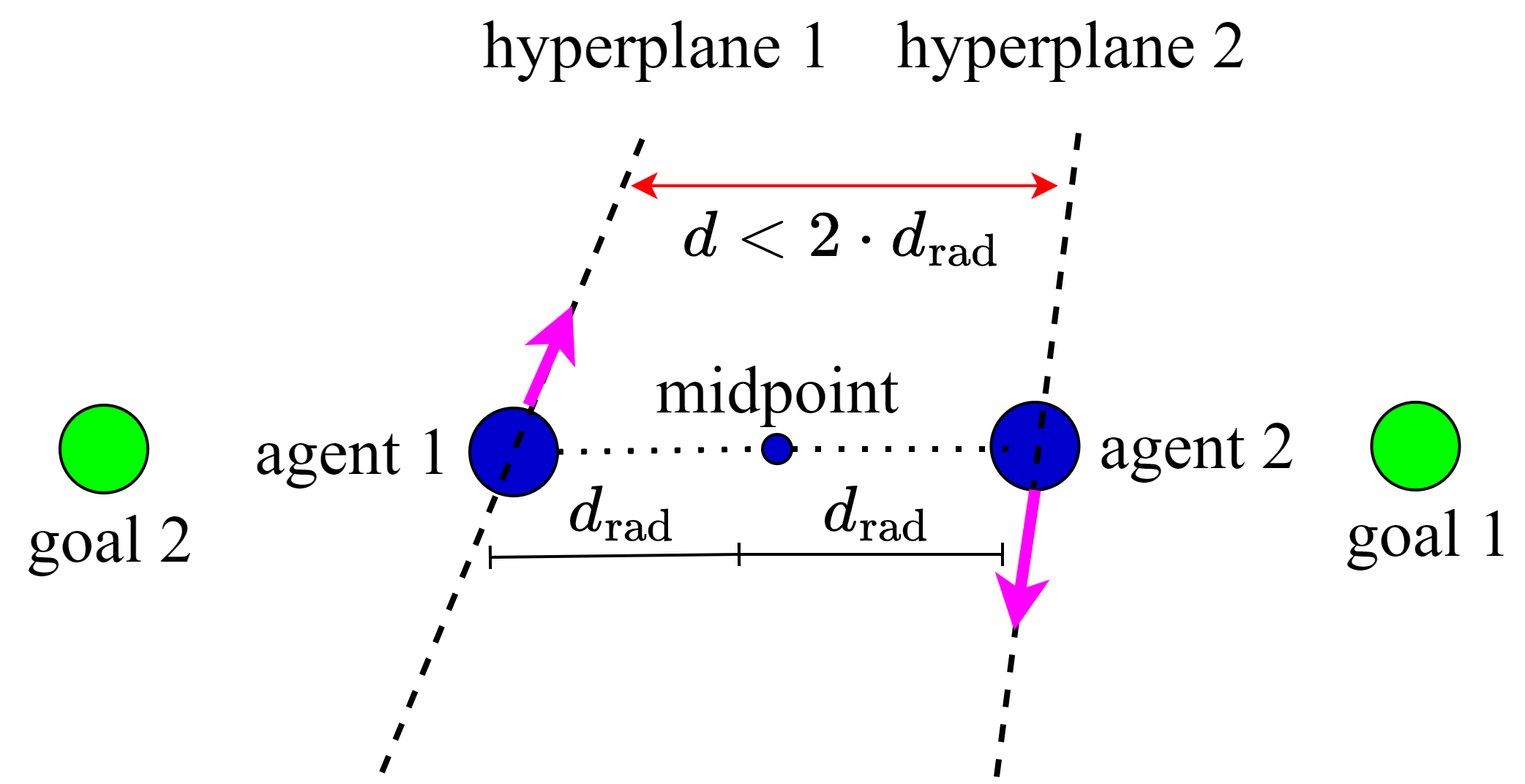}
\caption{Perturbing hyperplanes asymmetrically.}
\label{fig:stalemate_2}
\end{subfigure}
\begin{subfigure}{0.33\textwidth}
\centering
\includegraphics[trim={0cm 0cm 0cm 0cm},clip,width=0.9\linewidth]{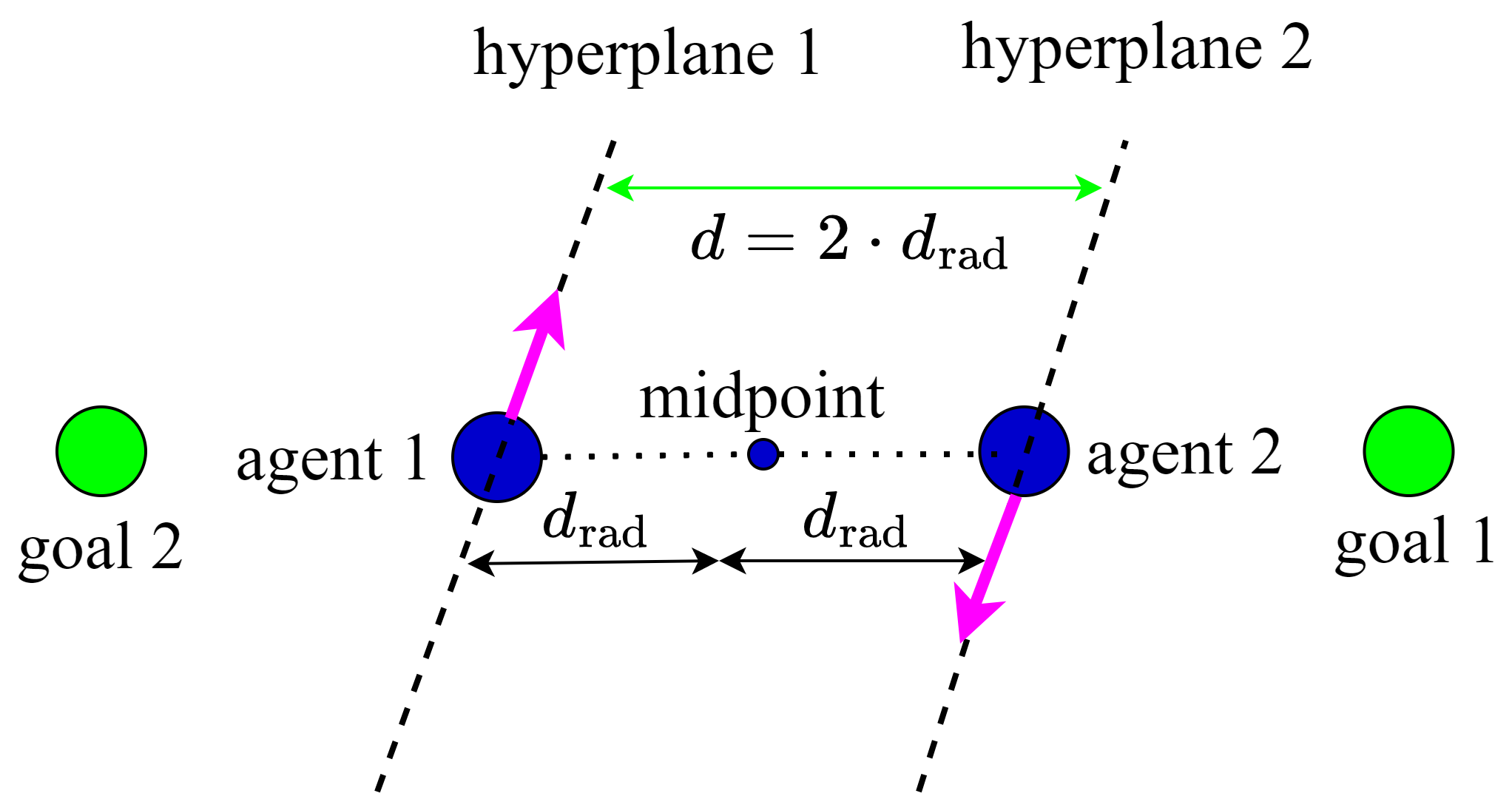}
\caption{Perturbing hyperplanes symmetrically.}
\label{fig:stalemate_3}
\end{subfigure}

\caption{A stalemate/deadlock happens when 2 agents are trying to move towards opposite goals and the solver is stuck on the borders of the hyperplanes (Fig. \ref{fig:stalemate_1}). Any movement up or down would not decrease the distance to the goal. If the hyperplanes are perturbed asymmetrically as done in \cite{toumieh2022multi} (Fig. \ref{fig:stalemate_2}), the distance between the agents can potentially become lower than the safety distance. We modify the perturbation vector (Sect. \ref{sect:ta_sc}) to make the perturbation symmetrical and guarantee safety when the agents move in the direction of the \textbf{magenta} vectors or any other direction (Fig. \ref{fig:stalemate_3}).}
\label{fig:stalemate}
\end{figure*}

\begin{figure}
\centering
\includegraphics[trim={-0.25cm -0.25 -0.25 -0.25cm},clip,width=1\linewidth]{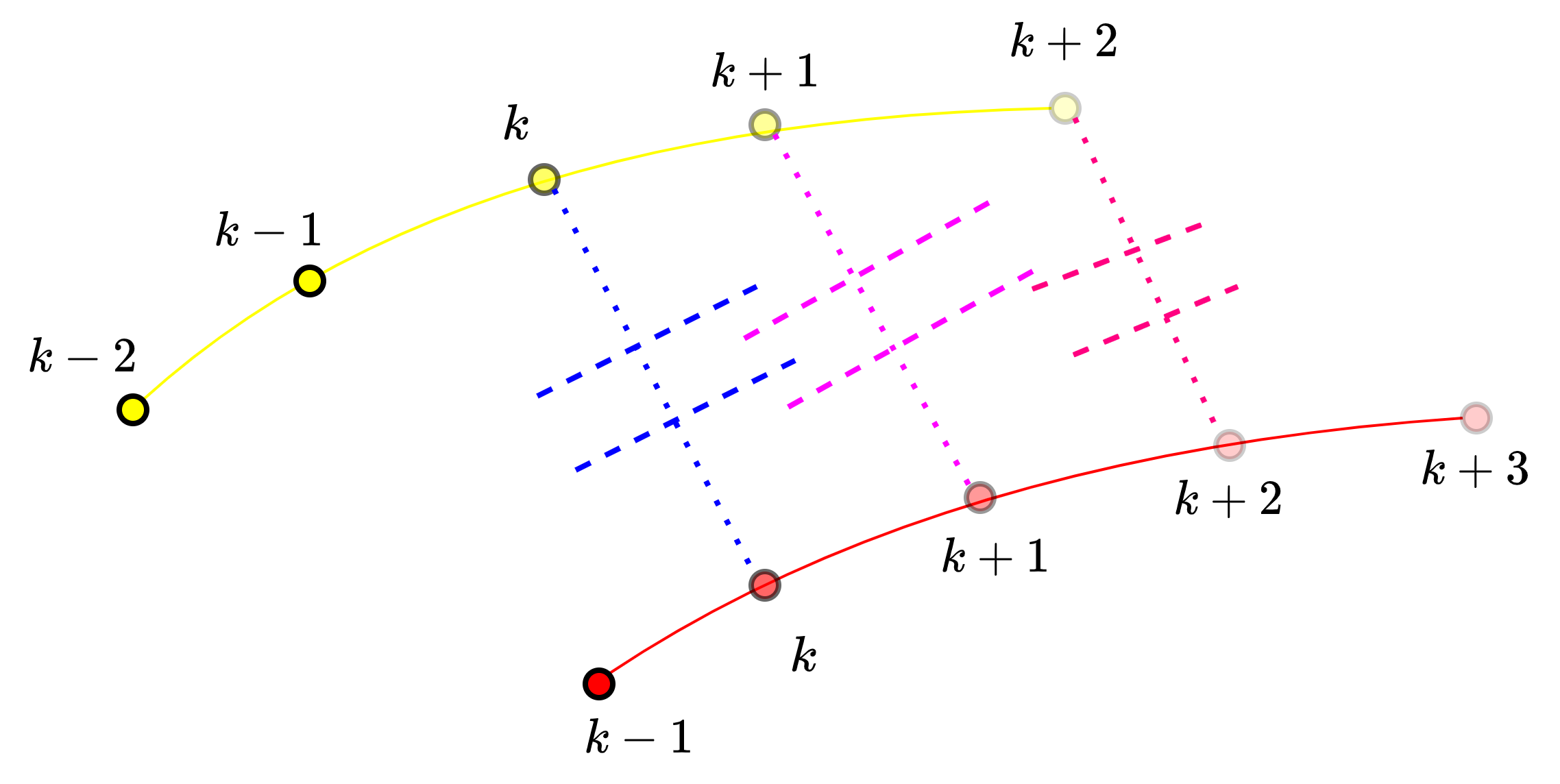}
\caption{We show the trajectories of 2 agents (in \textbf{red} and \textbf{yellow}) and the corresponding discrete positions that get more transparent as we move forward in time. We ignore the positions of each trajectory that have no corresponding position in the other ($k-2$ and $k+3$). The separating hyperplanes (\textbf{dashed lines} in different colors) are generated between the positions of the agents corresponding to the same time in the future starting from the current iteration $k$. The last separating hyperplane $k+2$ is used to fill the remaining $N-3$ hyperplanes required to generate the TASC.}
\label{fig:sep_planes}
\end{figure}
\subsection{Generate a Safe Corridor around the global path} \label{sect:sc}

Safe Corridors are a series of overlapping convex shapes that cover only free space in the environment. They are used by many state-of-the-art planning methods to constrain a dynamically feasible trajectory inside them, and thus guarantee safety \cite{toumieh2022near}, \cite{toumieh2020planning}, \cite{toumieh2022multi}. Many methods exist in the literature for Safe Corridor generation \cite{deits2015computing}, \cite{liu2017planning}, \cite{toumieh2020convex} \cite{toumieh2022shape}. The method used for the generation is \cite{toumieh2020convex} since it provides the best performance among the state-of-the-art methods for trajectory planning.

The Safe Corridor generation method takes as input a voxel grid (the local voxel grid centered around the agent) and the global path around which we want to generate the Safe Corridor. At each iteration, we always make sure that we have a certain number $P_{\text{hor}}$ of polyhedra that cover the free space of the environment.

At the first iteration of planning, we use the global path at the first iteration to generate a Safe Corridor that contains up to $P_{\text{hor}}$ number of polyhedra (polyhedra horizon). Subsequently, at each planning period, we use the global path generated in this planning period to update the Safe Corridor generated in the last step. The update consists of the following (Fig. \ref{fig:update_sc}): all the polyhedra that contain at least one point of the last generated MPC trajectory are kept. The other polyhedra are removed and new polyhedra are generated in their place until we have $P_{\text{hor}}$ polyhedra in total. To generate each polyhedron, we sample the global path at a constant step (voxel size). We then use the first point of the sampled global path that is outside all the remaining polyhedra as a seed voxel to generate an additional polyhedron.

\subsection{Generate a time-aware Safe Corridor (TASC)} \label{sect:ta_sc}
After generating the Safe Corridor, we use it along with the trajectories generated by all the other agents at the previous iterations to create a time-aware Safe Corridor (TASC). The future positions predicted by the MPC trajectories of the agents at the previous planning iterations are used to generate hyperplanes to constrain the future/MPC positions at the current iteration. These hyperplanes are added to the constraints of the Safe Corridor. This creates a series of Safe Corridors at each planning iteration that we call time-aware Safe Corridors in \cite{toumieh2022multi}. We refer the reader to \cite{toumieh2022multi} for a detailed explanation of how time-aware Safe Corridors are generated.

We augment/improve the TASC generation method to account for trajectories that were not generated at the same planning iteration $k$ (Fig. \ref{fig:sep_planes}). We ignore the positions of each trajectory that have no corresponding positions in the other trajectory ($k-1$ and $k+3$ in Fig. \ref{fig:sep_planes}). Then, starting with the position of the current iteration $k$, we generate separating hyperplanes for the rest of the common positions ($k$, $k+1$ and $k+2$ in Fig. \ref{fig:sep_planes}). Since we need $N$ separating hyperplanes to generate the TASC (as shown in \cite{toumieh2022multi}), we set the rest of the hyperplanes equal to the last separating hyperplanes ($k+2$ in Fig. \ref{fig:sep_planes}).
\subsubsection{Dealing with stalemates/deadlocks}
In \cite{toumieh2022multi}, in order to avoid stalemates/deadlocks, we modified the normal vectors of the separating hyperplanes by perturbing them constantly through time (a time-varying right-hand rule). This would avoid adding an explicit mechanism that creates subgoals for each agent to avoid stalemates/deadlocks like in \cite{Park2022DecentralizedDT}.
We defined the normalized plane normal $\boldsymbol{n}_{\text{hyp,norm}}$, the \textit{right} vector $\boldsymbol{r}$ that is the cross product between $\boldsymbol{n}_{\text{hyp,norm}}$ and $\boldsymbol{z}_W$ plus the cross product between $\boldsymbol{n}_{\text{hyp,norm}}$ and $\boldsymbol{y}_W$, a perturbation $m$, and a user-chosen coefficient $c$ that defines how tilted the final normal vector of the hyperplane $\boldsymbol{n}_{\text{hyp,final}}$ is with respect to the initial vector $\boldsymbol{n}_{\text{hyp}}$:
\begin{gather}
    \boldsymbol{n}_{\text{hyp,norm}} =    \dfrac{\boldsymbol{n}_{\text{hyp}}}{||\boldsymbol{n}_{\text{hyp}}||_2} \\
    \boldsymbol{z}_W = [0, 0, 1]^T, \quad
    \boldsymbol{y}_W = [0, 1, 0]^T \\ \boldsymbol{r} = \boldsymbol{n}_{\text{hyp,norm}} \times \boldsymbol{z}_W + \boldsymbol{n}_{\text{hyp,norm}} \times \boldsymbol{y}_W\\
    \boldsymbol{n}_{\text{pert}} = (c + m)\cdot \dfrac{\boldsymbol{r}}{||\boldsymbol{r}||_2} + c\cdot\boldsymbol{z}_W\\
    \boldsymbol{n}_{\text{hyp,final}} = \boldsymbol{n}_{\text{pert}} + \boldsymbol{n}_{\text{hyp,norm}} \label{eqn:pert}
\end{gather}

However, a component of the perturbation vector $\boldsymbol{n}_{\text{pert}}$ is non-symmetric ($c\cdot\boldsymbol{z}_W$), which can generate normal vectors that are non-colinear. This can result in cases where the distance between agents is lower than the safety/collision distance $2\cdot d_{\text{rad}}$ (Fig. \ref{fig:stalemate}). For this reason, we replace the non-symmetric term with the following symmetric term: $c\cdot(\boldsymbol{z}_W \times \boldsymbol{n}_{\text{hyp,norm}})$. The final perturbation vector then becomes:
\begin{gather}
    \boldsymbol{n}_{\text{pert}} = (c + m)\cdot \dfrac{\boldsymbol{r}}{||\boldsymbol{r}||_2} + c\cdot(\boldsymbol{z}_W \times \boldsymbol{n}_{\text{hyp,norm}})
\end{gather}

It is then added to $\boldsymbol{n}_{\text{hyp,norm}}$ to generate $\boldsymbol{n}_{\text{hyp,final}}$ as in equation (\ref{eqn:pert}).

\subsection{Generate a local reference trajectory} 
\label{sect:loc_ref}
We use the global path to generate a local reference trajectory that is used as a reference for the MPC to follow. The generation of such reference trajectory is done by sampling the global path at a constant velocity $v_{\text{samp}}$. The number of sampled points is equal to the number of discretization steps ($N$) in the MPC/MIQP formulation.

We only generate a new local reference trajectory in the following case: the last point of the MPC trajectory is within a distance $d_{\text{thresh}}$ from the last point of the local reference trajectory generated at the previous iteration. Otherwise, we keep the local reference trajectory generated at the previous planning iteration.

\begin{figure*}
\centering
\includegraphics[trim={0cm 0cm 0cm 0cm},clip,width=1\linewidth]{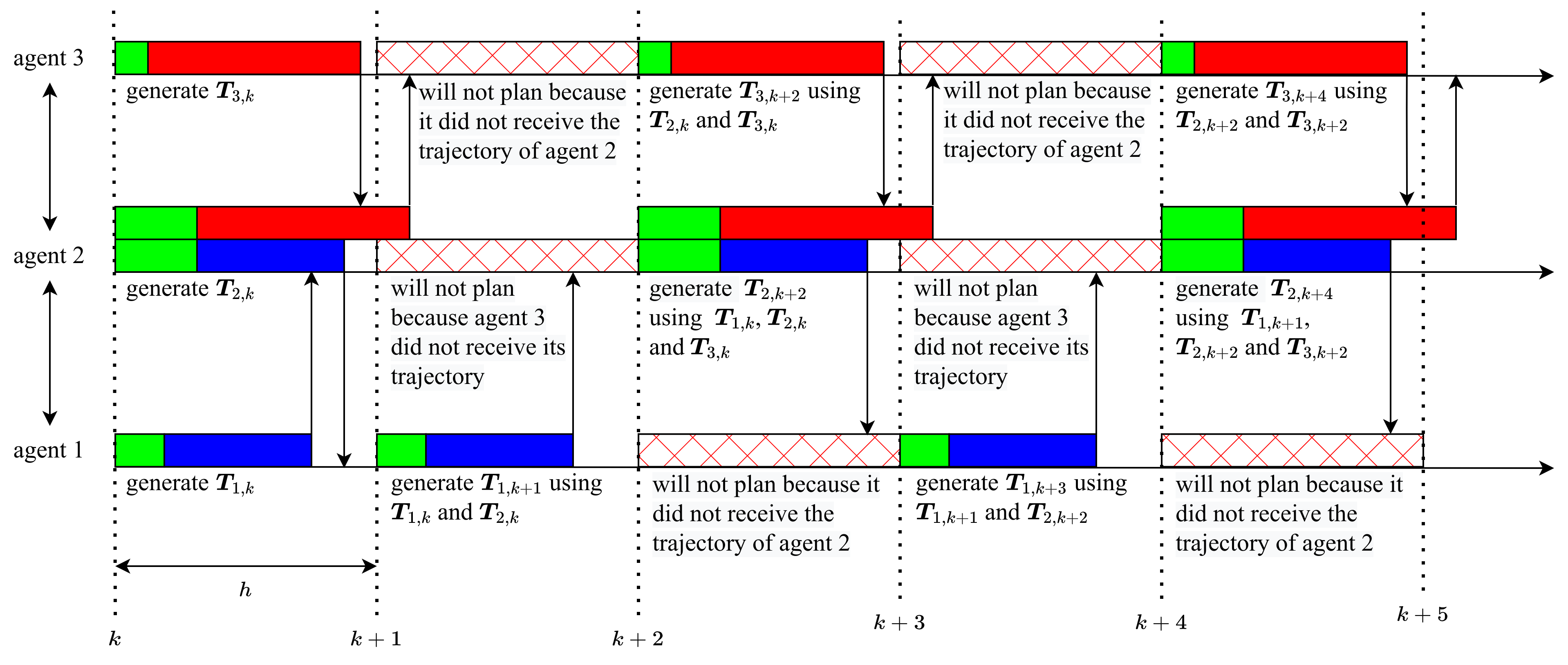}

\caption{We show an example of how different agents handle communication delays between each other. In this example agent 2 communicates with agents 1 and 3, whereas agents 1 and 3 do not communicate with each other (not within the range of communication). We show in \textbf{green} the computation time of each agent, in \textbf{blue} the communication latency between agents 1 and 2, and in \textbf{red} the communication latency between agents 2 and 3. The arrows indicate the time at which an agent $i$ receives the trajectory $\boldsymbol{T}_{j,k}$ of another agent $j$ generated at iteration $k$. At the first iteration, all agents synchronize their first planning iteration to be at the same time. At the subsequent iterations, an agent skips planning in one of 2 cases: 1) At least one agent within the communication range is yet to receive its last generated trajectory 2) It is yet to receive a new generated trajectory of another agent within the communication range and it has used all the previously received trajectories of this agent to generate its own trajectory.}
\label{fig:com_example}
\end{figure*}

\subsection{Solving the MIQP/MPC problem} \label{sect:form}
In this final step, we take the reference trajectory, and we solve an MPC optimization problem that minimizes the distance of the generated trajectory to the reference trajectory while also minimizing the jerk for smoothness. The generated trajectory consists of $N+1$ discrete states $\boldsymbol{x}_i$, $i = 0,1,...,N$ that contain the position, velocity, and acceleration of the agent. Each consecutive pair of discrete states are separated by a time step $h$. Thus, the time horizon of the planning is $N\cdot h$. The velocity and acceleration of the last state $\boldsymbol{x}_N$ are constrained/set to 0 to guarantee a safe trajectory for all agents in case subsequent optimizations fail (see \cite{toumieh2022multi} for more details).

The time-aware Safe Corridor is used to ensure the safety of the trajectory. We add the linear constraints of the time-aware Safe Corridor to the MPC optimization problem. By forcing each segment of the MPC trajectory be in at least one of the polyhedra of the time-aware Safe Corridor, we ensure no collision happens between the agent and the static obstacles as well as other planning agents. The final formulation of the optimization problem is a Mixed-Integer Quadratic Problem (MIQP) exactly like the one presented in \cite{toumieh2020planning}, \cite{toumieh2022multi}.

\subsection{Handling communication delay}
\label{sect:com_latency}

\begin{algorithm}
\caption{Run at every iteration $k$ for agent $i$:}\label{alg:latency}
\begin{algorithmic}[1]
\State delay\_planning = \textbf{false}
\For{each agent $j$ in $J$}
\If{received $\boldsymbol{T}_j$}
\State traj\_old[$j$].add($\boldsymbol{T}_j$)
\Else{}
\If{traj\_old[$j$].size() == 0}
\State delay\_planning = \textbf{true}
\EndIf
\EndIf
\If{not(delay\_planning)}
\State $dt_{\text{delay},i,j}$ = ComputeLatency(traj\_old[$j$][0])
\If{$dt_{\text{delay},i,j} + \boldsymbol{T}_{i,\text{last}}$.end} $> t_{\text{cur}}$
\State delay\_planning = \textbf{true}
\EndIf
\EndIf
\EndFor
\If{not(delay\_planning)}
\For{each agent $j$ in $J$}
\State GenerateTASC(traj\_old[$j$][0], $\boldsymbol{T}_{i,\text{last}}$)
\State traj\_old[$j$].RemoveFirstElement()
\EndFor
\EndIf
\end{algorithmic}
\end{algorithm}

Our previous work \cite{toumieh2022multi} ran the planning algorithm at a constant period equal to the MPC discretization step $dt_{\text{plan}} = h$. It was able to handle communication delay passively by assuming that the communication delay was lower than a time variable $dt_{\text{max,delay}}$ equal to the planning period $dt_{\text{plan}}$ minus the planner computation time $dt_{\text{comp}}$ ($dt_{\text{max,delay}} = dt_{\text{plan}} -  dt_{\text{comp}}$). However, no mechanism was in place to handle the communication latency when it exceeds $dt_{\text{max,delay}}$. 

In this work, we propose to adapt the planning period to be able to guarantee safety no matter the communication delay. 
 In addition to broadcasting the trajectory $\boldsymbol{T}_j$ when it finishes generating it, each agent $j$ broadcasts the time at which it started generating its trajectory i.e. the time at the start of the planning period ($\boldsymbol{T}_j$.start). It also broadcasts the time it finished generating the trajectory i.e. the time it sent it ($\boldsymbol{T}_j$.end). This allows another agent $i$ to estimate the communication delay between it and agent $j$ since their clocks are synchronized. The delay can be estimated by subtracting $\boldsymbol{T}_j$.end from the reception time of agent $i$, $t_{\text{rec},i}$:
 \begin{gather}
     dt_{\text{delay},i,j}= t_{\text{rec},i} - \boldsymbol{T}_j\text{.end} \label{eqn:delay}
 \end{gather}
 This in turn allows agent $i$ to know whether its last generated trajectory $\boldsymbol{T}_{i,\text{last}}$ was received by agent $j$ before the start time of the current planning period $t_{\text{cur}}$. The last generated trajectory of agent $i$ is not yet received by agent $j$ if the following condition is true:
 \begin{gather}
 dt_{\text{delay},i,j} +  \boldsymbol{T}_{i,\text{last}}.\text{end} >  t_{\text{cur}}  \label{eqn:delay_check}
\end{gather}
 The planner will skip planning at the start of the current planning period and wait for the next period if one of these 2 cases is true:
\begin{enumerate}
    \item It knows that there is another agent within its communication range that is yet to receive its last planned trajectory.
    \item It is yet to receive a new planned trajectory of another agent within its communication range and it has used all the old received trajectories of this agent for planning.
\end{enumerate}
 
 We propose the following algorithm to handle communication latency (Alg. \ref{alg:latency}).
  At every planning iteration (which happens every $dt_{\text{plan}} = h$), every agent $i$ checks if it received a trajectory from every other agent $j$ (line 3). If it did, it adds the received trajectory to a 2D vector (traj\_old) whose first index indicates the number or ID of the other agent i.e. $j$ (line 4). If agent $i$ did not receive a trajectory from agent $j$, it checks if there is an unused old trajectory in the vector traj\_old[$j$] (line 5-6). If not, we delay the planning since we have no new or old trajectory to use for generating the TASC (line 7). If the planning should not be delayed due to previous conditions (line 8), we check if it should be delayed because agent $j$ hasn't received the trajectory of agent $i$ yet. This is done by first computing the communication delay using equation (\ref{eqn:delay}) (line 9), and then checking the condition (\ref{eqn:delay_check}) (lines 10-11). Finally, we check if the planning should be delayed after going through all agents (line 12). If not, we compute the TASC using the oldest unused trajectory of each agent $j$ and remove it from the vector of old trajectories (lines 13-15). The starting time $\boldsymbol{T}_j$.start allows to know at which iteration $k$ the trajectory was generated, which is important in TASC generation (Fig. \ref{fig:sep_planes}).
  
  We show an example of how this algorithm would perform in Fig. \ref{fig:com_example}. In this example, agent 2 sees and communicates with agents 1 and 3, but agents 1 and 3 do not see and communicate with each other. Still, the algorithm allows for safe planning and coordination between all agents.

\section{Simulation Results} \label{sect:sim_res}
\begin{figure*}
\begin{subfigure}{0.32\textwidth}
\centering
\includegraphics[trim={1cm 0cm 1cm 0cm},clip,width=1.05\linewidth]{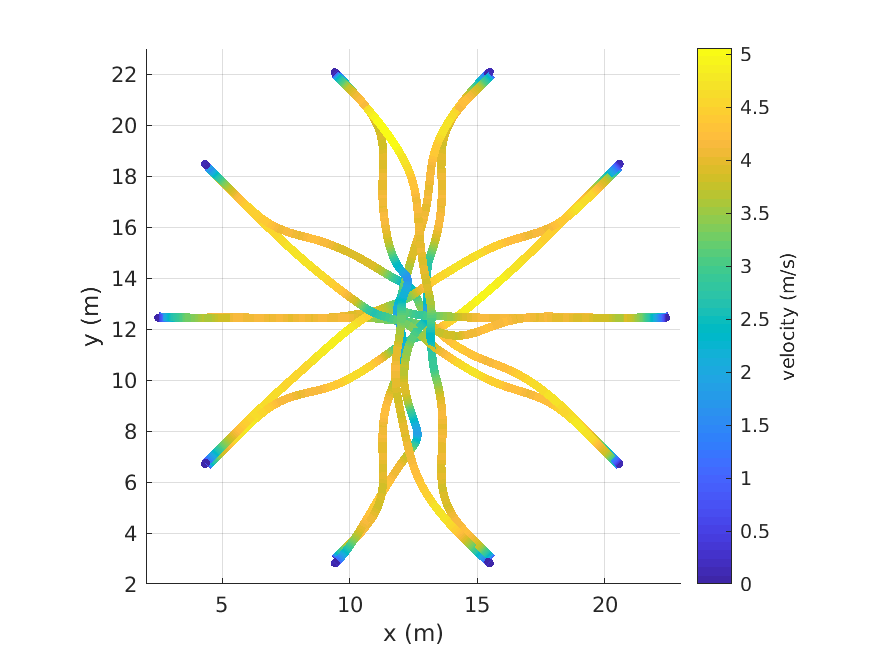}
\caption{Our planner: 10 agents with $dt = 100$ ms with the setup in Tab. \ref{table:comparison_table_mader}.}
\label{fig:our_100}
\end{subfigure}\hfill%
\begin{subfigure}{0.32\textwidth}
\centering
\includegraphics[trim={1cm 0cm 1cm 0cm},clip,width=1.05\linewidth]{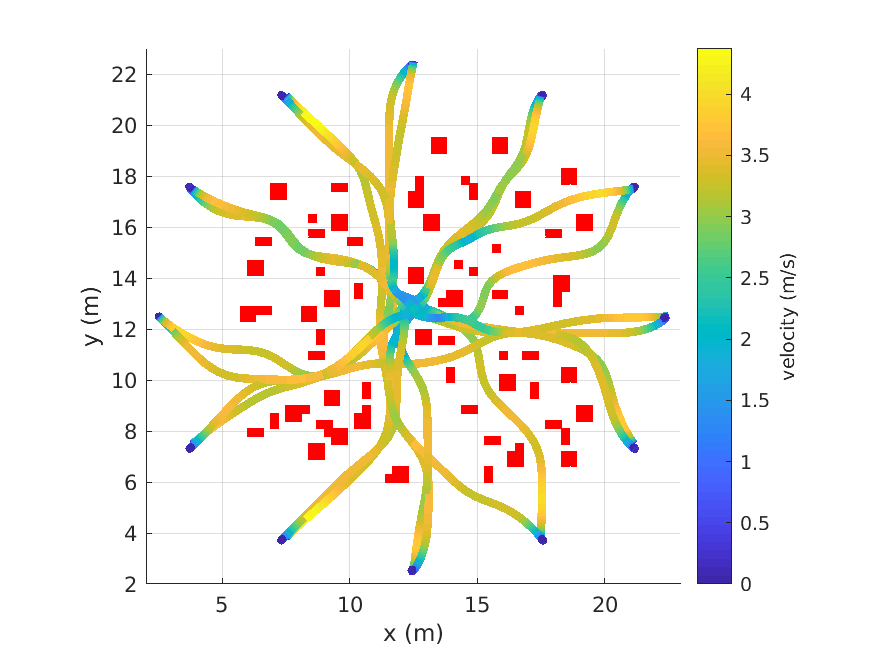}
\caption{Our planner: 12 agents with $dt = 0$ ms and obstacles (Sect. \ref{sect:sim_obs}).}
\label{fig:our_obs_0}
\end{subfigure}\hfill%
\begin{subfigure}{0.32\textwidth}
\centering
\includegraphics[trim={1cm 0cm 1cm 0cm},clip,width=1.05\linewidth]{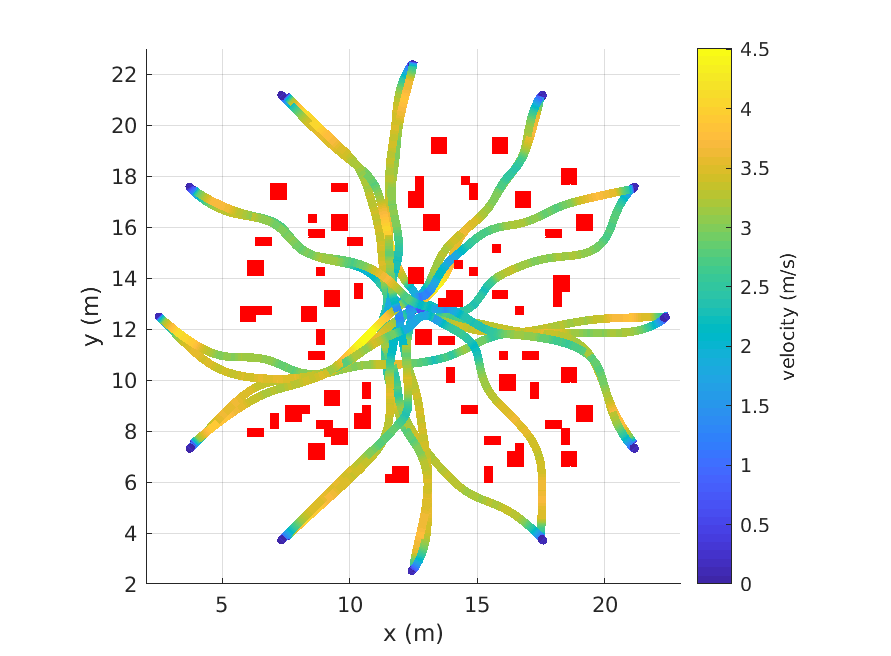}
\caption{Our planner: 12 agents with $dt = 150$ ms and obstacles (Sect. \ref{sect:sim_obs}).}
\label{fig:our_obs_150}
\end{subfigure}
\caption{The agents start in a circular configuration and swap positions. We show an overhead view of the trajectories generated by our planner in different settings (with and without obstacles), different communication latencies, and different dynamic limits.}
\label{fig:our_dt}
\end{figure*}

    The testing setup is similar to what is presented in \cite{kondo2022robust}. Thus, we will use their results as a reference for our comparison. The simulations are run on Intel i7 CPUs with a base frequency of 2.6GHz and a turbo boost of 4GHz. The testing consists of 10 agents in a circular configuration (Fig. \ref{fig:our_100}) exchanging positions. We compare our method with RMADER \cite{kondo2022robust} and 2 versions of Ego-Swarm \cite{Zhou2021EGOSwarmAF}. We set the maximum velocity $v_{\text{max}} = 10\ \text{m/s}$, the maximum acceleration $a_{\text{max}} = 20\ \text{m/s\textsuperscript{2}}$ and the maximum jerk $j_{\text{max}} = 30\ \text{m/s\textsuperscript{3}}$ for RMADER, Ego-Swarm and our method (along the $x$, $y$ and $z$ directions). For Ego-Swarm, we also consider a more conservative version (slow Ego-Swarm) with a maximum acceleration $a_{\text{max}} = 10\ \text{m/s\textsuperscript{2}}$ and a maximum velocity $v_{\text{max}} = 5\ \text{m/s}$. 
    
    For MADER and RMADER, each agent is represented as a bounding box of size $0.25\times0.25\times0.25$ m. For Ego-Swarm and our planner, each agent is represented as a sphere of diameter $0.25$ m as per the experiments in \cite{kondo2022robust} (at the time of writing, the bounding box dimensions and sphere diameter were not mentioned in \cite{kondo2022robust}, but they were communicated to us by the authors of \cite{kondo2022robust}).
    The comparison is done with 100 simulated runs for communication latencies equal to $0$, $50$, and $100$ milliseconds. The comparison metrics are:
\begin{enumerate}
\item Collision \%: percentage of simulations where there was at least one collision.
\item Average number of stops expected in a single simulation from all agents.
\item Mean of the jerk cost $J_{\text{cost}} = \int_{t_{\text{ini}}}^{t_{\text{fin}}} ||\boldsymbol{j}(t)||^2\mathrm{d}t$ where $t_{\text{ini}}$ and $t_{\text{fin}}$ are the initial and final time of the trajectory.
\item Mean of the acceleration cost $A_{\text{cost}} = \int_{t_{\text{ini}}}^{t_{\text{fin}}} ||\boldsymbol{a}(t)||^2\mathrm{d}t$.
\item Mean and max flight time.
\item Computation time.
\end{enumerate}

\begin{table*}[ht]
\centering
\caption{Comparison between Ego-Swarm (ES) \cite{Zhou2021EGOSwarmAF}, slow Ego-Swarm (Slow ES) \cite{Zhou2021EGOSwarmAF}, MADER \cite{tordesillas2020mader}, RMADER \cite{kondo2022robust} and our method. The comparison consists of 100 simulations with communication delays between 10 agents exchanging positions in a circular configuration as in Fig. \ref{fig:our_100}. The communication delays are \textcolor{blue}{$dt = 0$ ms} $\mid$ \textcolor{OliveGreen}{$dt = 50$ ms} $\mid$ \textcolor{magenta}{$dt = 100$ ms}. We show in bold the best performer among the safe planners (RMADER \cite{kondo2022robust} and our planner).}
\begin{tabular}{ c | ccccccc}\hline
 Method & Collision [\%] & Mean \# stops & Accel. cost (m/s\textsuperscript{2}) & Jerk cost (10\textsuperscript{3} m/s\textsuperscript{3}) & Mean flight time (s) & Max flight time (s) \\ \hhline{=======}
ES \cite{Zhou2021EGOSwarmAF} & \textcolor{blue}{64} $\mid$ \textcolor{OliveGreen}{84} $\mid$ \textcolor{magenta}{84} & \textcolor{blue}{0.004} $\mid$ \textcolor{OliveGreen}{0} $\mid$ \textcolor{magenta}{0.01} & \textcolor{blue}{662} $\mid$ \textcolor{OliveGreen}{700} $\mid$ \textcolor{magenta}{788} & \textcolor{blue}{9.07} $\mid$ \textcolor{OliveGreen}{9.46} $\mid$ \textcolor{magenta}{10.4} & \textcolor{blue}{7.19} $\mid$ \textcolor{OliveGreen}{7.24} $\mid$ \textcolor{magenta}{7.28} & \textcolor{blue}{7.38} $\mid$ \textcolor{OliveGreen}{7.51} $\mid$ \textcolor{magenta}{7.63}\\ \hline
Slow ES \cite{Zhou2021EGOSwarmAF} & \textcolor{blue}{14} $\mid$ \textcolor{OliveGreen}{25} $\mid$ \textcolor{magenta}{22} & \textcolor{blue}{0} $\mid$ \textcolor{OliveGreen}{0} $\mid$ \textcolor{magenta}{0} & \textcolor{blue}{110} $\mid$ \textcolor{OliveGreen}{113} $\mid$ \textcolor{magenta}{113} & \textcolor{blue}{15.4} $\mid$ \textcolor{OliveGreen}{15.5} $\mid$ \textcolor{magenta}{15.5} & \textcolor{blue}{11.6} $\mid$ \textcolor{OliveGreen}{11.7} $\mid$ \textcolor{magenta}{11.8} & \textcolor{blue}{11.9} $\mid$ \textcolor{OliveGreen}{12} $\mid$ \textcolor{magenta}{13}\\ \hline
MADER \cite{tordesillas2020mader} & \textcolor{blue}{15} $\mid$ \textcolor{OliveGreen}{38} $\mid$ \textcolor{magenta}{42} & \textcolor{blue}{0} $\mid$ \textcolor{OliveGreen}{0.001} $\mid$ \textcolor{magenta}{0} & \textcolor{blue}{78.1} $\mid$ \textcolor{OliveGreen}{74.2} $\mid$ \textcolor{magenta}{74.5} & \textcolor{blue}{1.59} $\mid$ \textcolor{OliveGreen}{1.64} $\mid$ \textcolor{magenta}{1.64} & \textcolor{blue}{6.28} $\mid$ \textcolor{OliveGreen}{6.25} $\mid$ \textcolor{magenta}{6.26} & \textcolor{blue}{7.15} $\mid$ \textcolor{OliveGreen}{7.35} $\mid$ \textcolor{magenta}{7.04}\\ \hline
RMADER \cite{kondo2022robust} & \textcolor{blue}{0} $\mid$ \textcolor{OliveGreen}{0} $\mid$ \textcolor{magenta}{0} & \textcolor{blue}{0.46} $\mid$ \textcolor{OliveGreen}{0.347} $\mid$ \textcolor{magenta}{1.75} & \textcolor{blue}{127} $\mid$ \textcolor{OliveGreen}{148} $\mid$ \textcolor{magenta}{190} & \textcolor{blue}{2.94} $\mid$ \textcolor{OliveGreen}{3.71} $\mid$ \textcolor{magenta}{5.94} & \textcolor{blue}{7.28} $\mid$ \textcolor{OliveGreen}{7.95} $\mid$ \textcolor{magenta}{10.4} & \textcolor{blue}{8.41} $\mid$ \textcolor{OliveGreen}{8.80} $\mid$ \textcolor{magenta}{11.9}\\ \hline
proposed & \textcolor{blue}{0} $\mid$ \textcolor{OliveGreen}{0} $\mid$ \textcolor{magenta}{0} & \textcolor{blue}{\textbf{0}} $\mid$ \textcolor{OliveGreen}{\textbf{0}} $\mid$ \textcolor{magenta}{\textbf{0}} & \textcolor{blue}{\textbf{109}} $\mid$ \textcolor{OliveGreen}{\textbf{114}} $\mid$ \textcolor{magenta}{\textbf{119}} & \textcolor{blue}{\textbf{2.27}} $\mid$ \textcolor{OliveGreen}{\textbf{2.49}} $\mid$ \textcolor{magenta}{\textbf{5.03}} & \textcolor{blue}{\textbf{6.77}} $\mid$ \textcolor{OliveGreen}{\textbf{6.79}} $\mid$ \textcolor{magenta}{\textbf{7.1}} & \textcolor{blue}{\textbf{7.1}} $\mid$ \textcolor{OliveGreen}{\textbf{7.3}} $\mid$ \textcolor{magenta}{\textbf{7.7}}\\ \hline
\end{tabular}
\label{table:comparison_table_mader}
\end{table*}

\begin{table}[ht]
\centering
\caption{Computation time of our planner for the results in Tab. \ref{table:comparison_table_mader}. We show the \textbf{mean / max / standard deviation}.}
\begin{tabular}{c|c|c|c}\hline
& $dt = 0$ ms & $dt = 50$ ms & $dt = 100$ ms \\
\hhline{====}
Comp. (ms) & 10.4 / 61 / 6.6 & 10.1 / 54.7 / 6.4 & 11.4 / 70 / 6.7 \\ \hline
\end{tabular}
\label{table:comp_time_our}
\end{table}

\subsection{Planner parameters}
The local voxel grid around each agent is of size $15\times 15\times 3.3$ m and has a voxel size of $0.3$ m.  We choose the following parameters: $N = 9$, $h = 100$ ms, $v_{\text{samp}} = 4.5$ m/s, $P_{\text{hor}} = 3$, $d_{\text{thresh}}= 0.4$ m. 
The rest of the parameters are chosen the same as in \cite{toumieh2022multi} with the exception of the maximum velocity, acceleration, and jerk which are the same for all planners (Sect. \ref{sect:sim_res}).

\subsection{Comparison with the state-of-the-art}
We show in Tab. \ref{table:comparison_table_mader} the results of the planners with different communication latencies ($0$, $50$, and $100$ ms). Our planner and Ego-Swarm \cite{Zhou2021EGOSwarmAF} use voxel girds as representations of the obstacles in the environment. MADER \cite{tordesillas2020mader} and RMADER \cite{kondo2022robust} on the other hand use a polyhedral representation of the environment i.e. all obstacles are represented by a series of convex polyhedra. This representation is not trivial to generate and may add considerable overhead to the autonomous navigation pipeline. 

Our planner and RMADER \cite{kondo2022robust} are the only planners that are able to generate collision-free trajectories in all simulations, so we will focus our comparison on them. Our planner outperforms RMADER in trajectory smoothness across all latencies using both the acceleration ($25$\% better on average) and the jerk ($24$\% better on average) metrics.

The mean and max flight times of our planner grow slower than those of RMADER with the increase in latency. Over all latencies, our planner outperforms RMADER in mean flight time by an average of $18$\% and max flight time by an average of $23$\%.

\subsubsection{Computation time} \label{sect:comp_time}
Ego-Swarm is the most computationally efficient with an average computation time of $0.5$ ms. RMADER improves on MADER \cite{tordesillas2020mader} in computation time by changing the optimization problem from non-convex to convex. This improves the mean computation time by $20$\% (from $39.23$ ms to $31.08$ ms) and the max computation time by $40$\% (from $724$ ms to $433$ ms) as reported in \cite{kondo2022robust}. While our planner is not as efficient as Ego-Swarm, it is much more efficient than RMADER as shown in Tab. \ref{table:comp_time_our}. The mean computation time across all latencies is $10.6$ ms and the max is $70$ ms. 

\begin{table*}[ht]
\centering
\caption{Results for 8 and 12 agents in an environment with obstacles (Sect. \ref{sect:sim_obs}). The \textbf{mean / max / standard deviation} of each metric is shown.}
\begin{tabular}{c|c|ccccccc}\hline
 \# & $dt$ (ms) & Distance (m) & Velocity (m/s) & Flight time (s) & Comp. time (ms) & Acc. cost (m/s\textsuperscript{2}) & Jerk cost (10\textsuperscript{3}m/s\textsuperscript{3}) \\ \hhline{========}
 \multirow{4}{*}{8} & 0 & 21.6 / 23.1 / 0.72 & 2.52 / 4.21 / 1.24 & 8.47 / 9.5 / 0.4 & 5.5 / 48.7 / 3 & 121 / 170 / 26.2 & 3.5 / 5.56 / 0.94 \\ \cline{2-8}
 & 50 & 21.6 / 23.1 / 0.72 & 2.51 / 4.21 / 1.24 & 8.47 / 9.5 / 0.4 & 5.4 / 48.1 / 2.9 & 121 / 170 / 26.2 & 3.5 / 5.56 / 0.95 \\ \cline{2-8}
 & 100 & 21.6/ 23.4 / 0.76 & 2.43 / 4.24 / 1.22 & 8.7 / 9.5 / 0.42 & 6.2 / 35 / 3.8 & 124 / 182 / 26.7 & 6.59 / 9.11 / 0.96 \\ \cline{2-8}
 & 150 & 21.6 / 23.4 / 0.76 & 2.43 / 4.24 / 1.22 & 8.7 / 9.5 / 0.42 & 6.1 / 33.3 / 3.8 & 124 / 182 / 26.5 & 6.59 / 9.11 / 0.96 \\ \hhline{========}
 \multirow{4}{*}{12} & 0 & 21.7 / 24.2 / 0.73 & 2.45 / 4.5 / 1.23 & 8.7 / 9.9 / 0.45 & 8.7 / 72.4 / 6 & 130 / 207 / 26.8 & 3.76 / 5.65 / 0.84 \\ \cline{2-8}
 & 50 & 21.7 / 24.1 / 0.73 & 2.46 / 4.5 / 1.24 & 8.7 / 9.9 / 0.43 & 8.4 / 69.6 / 5.8 & 136 / 207 / 27.6 & 4.19 / 6.56 / 0.88 \\ \cline{2-8}
 & 100 & 21.6 / 23.9 / 0.71 & 2.38 / 4.36 / 1.2 & 8.98 / 10.3 / 0.46 & 9.2 / 85.9 / 7.3 & 134 / 240 / 28.7 & 6.86 / 10.8 / 0.97 \\ \cline{2-8}
 & 150 & 21.7 / 23.7 / 0.7 & 2.36 / 4.86 / 1.22 & 9.08 / 10.4 / 0.44 & 10.8 / 86.6 / 8.4 & 146 / 308 / 34.2 & 8.41 / 17.1 / 1.46 \\ \cline{1-8}
 \end{tabular}
\label{table:comp_obs}
\end{table*}

\subsection{Environment with obstacles} \label{sect:sim_obs}
We add obstacles to the environment as well as delay to see how our planner performs as the communication latency increases. The obstacles have already been inflated by the agent's radius at their generation. We test for $8$ and $12$ agents. Furthermore, we change the diameter of each agent to $0.3$ m, $v_{\text{samp}} = 3.5$ m/s, $a_{\text{max}} = 30$ m/s\textsuperscript{2}, $j_{\text{max}} = 60$ m/s\textsuperscript{3}, $N = 7$ and $d_{\text{thresh}} = 0.2$ m for experimental diversity. We generate $70$ obstacles of size $0.2\times0.2\times1.5$ m with random positions at each simulation run (uniform distribution - Fig. \ref{fig:our_obs_0}, \ref{fig:our_obs_150}). We do 10 simulation runs for each latency $dt = 0, 50, 100$, and $150$ ms. The performance metrics used are the distance traversed by each agent, the flight velocity and time, the computation time, and the acceleration and jerk costs. The \textbf{mean / max / standard deviation} of each metric are shown in Tab. \ref{table:comp_obs}.

In all test runs for 8 and 12 agents, all agents were able to reach their intended goal/destination safely i.e. the safety distance between the agents was not violated and they did not get stuck along the way.

For 8 agents, the results for $dt = 0$ ms and $dt = 50$ ms are similar. This is due to the fact that in both cases, all agents receive the trajectories before the start of the next planning iteration since the maximum computation time is below $50$ ms. The results for $dt = 100$ ms and $dt = 150$ ms are also similar due to the same reason: in both cases, all agents receive the trajectories of other agents every 2 planning iterations (the planning period is effectively $2h$ due to our latency handling algorithm \ref{alg:latency}).

For 8 and 12 agents, the jerk cost and computation time both increase as the latency increases. This is due to the more frequent slowdown of each agent as the latency increases. The slowdown is due to passing through narrow spaces and avoiding other agents at the same time as well as the latency handling mechanism (see video link after the abstract).

\section{Conclusions and Future Works}
In this paper, we presented an improved decentralized, real-time, and synchronous framework for multi-agent planning. The method improves on our previous work \cite{toumieh2022multi} by making it fully online and suitable for real-world applications (the global path planning and Safe Corridor generation steps were done offline in \cite{toumieh2022multi}). Furthermore, we added a mechanism to handle arbitrary communication latency and adapt the planning frequency accordingly. Our previous work was only able to handle communication latency when it is lower than a predetermined threshold. We compared our work to 3 state-of-the-art multi-agent planning methods: Ego-Swarm \cite{Zhou2021EGOSwarmAF}, MADER \cite{tordesillas2020mader} and RMADER \cite{kondo2022robust}. We showed that our planner generates the safest trajectories with a $0$\% collision rate. Furthermore, it generates smoother and faster trajectories than the only other safe and latency robust planner (RMADER) while also being at least $3\times$ more computationally efficient.

In the future, we plan on implementing our planning method on embedded drone systems for swarm autonomous navigation. This would require implementing relative localization algorithms between agents, obstacle detection for collision avoidance, as well as a communication mechanism for broadcasting information between agents. Finally, we intend on developing a formation flight version of our planner. This can be done by adding a cost to the objective function of our planner that makes agents preserve a predefined shape.







\bibliographystyle{IEEEtran}
\bibliography{IEEEabrv,IEEEexample}

\end{document}